\documentclass[11pt,letterpaper]{article}
\usepackage{emnlp2017}
\usepackage{times}
\usepackage{latexsym}

\usepackage{graphicx}
\usepackage{booktabs}
\usepackage{tabularx}
\usepackage{amsmath}
\usepackage{todonotes}

\usepackage{color}

\emnlpfinalcopy

\def\emnlppaperid{***}

\title{Grasping the Finer Point:\\ A Supervised Similarity Network for Metaphor Detection}

\author{Marek Rei$^{\clubsuit\spadesuit}$ ~ Luana Bulat$^{\clubsuit}$ ~ Douwe Kiela$^{\diamondsuit}$ ~ Ekaterina Shutova$^{\clubsuit}$\\
$^\clubsuit$Computer Laboratory, University of Cambridge, United Kingdom \\
$^\spadesuit$The ALTA Institute, University of Cambridge, United Kingdom \\
$^\diamondsuit$Facebook AI Research, New York, USA \\
{ \small \tt \{marek.rei,luana.bulat,ekaterina.shutova\}@cl.cam.ac.uk, dkiela@fb.com}
}
\date{}

\date{}

\begin{document}

\maketitle

\begin{abstract}
The ubiquity of metaphor in our everyday communication makes it an important problem for natural language understanding. Yet, the majority of metaphor processing systems to date rely on hand-engineered features and there is still no consensus in the field as to which features are optimal for this task. In this paper, we present the first deep learning architecture designed to capture metaphorical composition. Our results demonstrate that it outperforms the existing approaches in the metaphor identification task.
\end{abstract}

\section{Introduction}

Metaphor is pervasive in our everyday communication, enriching it with sophisticated imagery and helping us to reconcile our experience in the world with our conceptual system \cite{LakoffAndJohnson}. In the most influential account of metaphor to date, Lakoff and Johnson explain the phenomenon through the presence of systematic metaphorical associations between two distinct concepts or domains. For instance, when we talk about ``\textit{curing} juvenile delinquency'' or ``corruption \textit{transmitting} through the government ranks'', we view the general concept of \textit{crime} (the target concept) in terms of the properties of a \textit{disease} (the source concept). Such metaphorical associations are broad generalisations that allow us to project knowledge and inferences across domains; and our metaphorical use of language is a reflection of this process.

Given its ubiquity, metaphorical language poses an important problem for natural language understanding \cite{Cameron2003,ShutovaLREC}. 
A number of approaches to metaphor processing have thus been proposed, focusing predominantly on classifying linguistic expressions as literal or metaphorical. They experimented with a range of features, including lexical and syntactic information \cite{hovy-EtAl:2013:Meta4NLP,BeigmanKlebanovACL2016} and higher-level features such as semantic roles \cite{Catching}, domain types \cite{DunnCicling2013}, concreteness \cite{Turney2011}, imageability \cite{strzalkowski-EtAl:2013:Meta4NLP} and WordNet supersenses \cite{Boytsov2014}. While reporting promising results, all of these approaches used hand-engineered features and relied on manually-annotated resources to extract them. In order to reduce the reliance on manual annotation, other researchers experimented with sparse distributional features \cite{ShutovaColing2010,ShutovaNAACL2013} and dense neural word embeddings \cite{Mohler2014Journal,Shutova2016}. Their experiments have demonstrated that corpus-driven lexical representations already encode information about semantic domains needed to learn the patterns of metaphor usage from linguistic data. 

We take this intuition a step further and present the first deep learning architecture designed to capture metaphorical composition. Deep learning methods have already been shown successful in many other semantic tasks \cite[e.g.][]{Hermann2015,Kumar2015,Zhao2015}, which suggests that designing a specialised neural network architecture for metaphor detection will lead to improved performance. In this paper, we present a novel architecture which (1) models the interaction between the source and target domains in the metaphor via a gating function; (2) specialises word representations for the metaphor identification task via supervised training; (3) quantifies metaphoricity via a weighted similarity function that automatically selects the relevant dimensions of similarity. We experimented with two types of word representations as inputs to the network: the standard skip-gram word embeddings \cite{Mikolov2013a} and the cognitively-driven attribute-based vectors \cite{Bulat2017}, as well as a combination thereof.

We evaluate our method in the metaphor identification task, focusing on adjective--noun, verb--subject and verb--direct object constructions where the verbs and adjectives can be used metaphorically. Our results show that our architecture outperforms both a metaphor agnostic deep learning baseline (a basic feed forward network) and the previous corpus-based approaches to metaphor identification. We also investigate the effects of training data on this task, and demonstrate that with a sufficiently large training set our method also outperforms the best existing systems based on hand-coded lexical knowledge.

\begin{figure*}[t]
    \centering
	\includegraphics[width=0.8\linewidth]{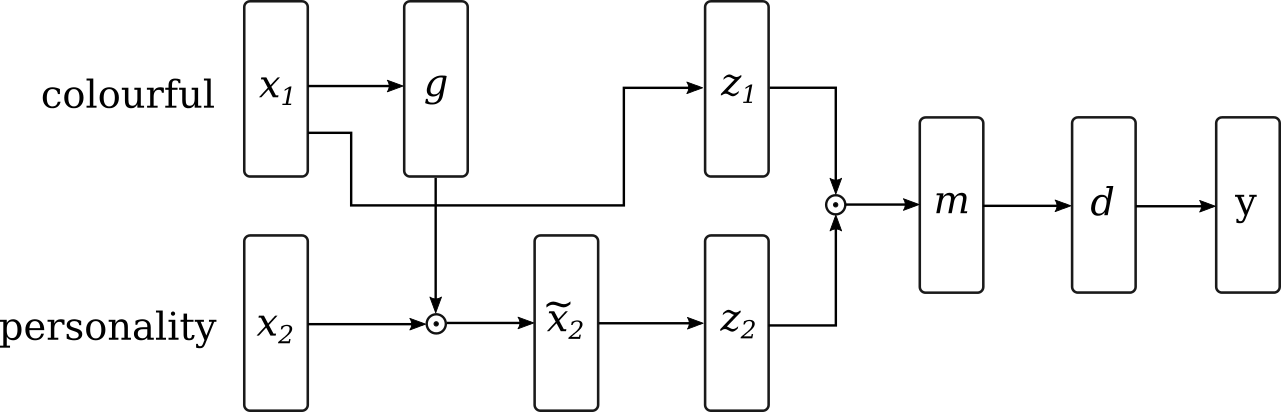}
	\caption{The network architecture for supervised metaphorical phrase classification. The $\odot$ symbol is used to indicate element-wise multiplication.}
	\label{fig:network}
\end{figure*}

\section{Related Work}

The majority of approaches to metaphor processing cast the problem as classification of linguistic expressions as metaphorical or literal. 
 \newcite{Catching} classified verbs related to \textsc{motion} and \textsc{cure} within the domain of financial discourse. They used the maximum entropy classifier and the verbs' nominal arguments and their FrameNet roles \cite{FrameNet} as features, reporting encouraging results. Dunn \shortcite{DunnCicling2013} used a logistic regression classifier and high-level properties of concepts extracted from SUMO ontology, including domain types (\textsc{abstract, physical, social, mental}) and event status (\textsc{process, state, object}). \newcite{Boytsov2014} used random forest classifier and coarse semantic features, such as concreteness, animateness, named entity types and WordNet supersenses. They have shown that the model learned with such coarse semantic features is portable across languages. 
  The work of \newcite{hovy-EtAl:2013:Meta4NLP} is notable as they focused on compositional rather than categorical features. They trained an SVM with dependency-tree kernels to capture compositional information, using lexical, part-of-speech tag and WordNet supersense representations of sentence trees. \newcite{mohler-EtAl:2013:Meta4NLP} aimed at modelling conceptual information. They derived semantic signatures of texts as sets of highly-related and interlinked WordNet synsets. The semantic signatures served as features to train a set of classifiers (maximum entropy, decision trees, SVM, random forest) that mapped new metaphors to the semantic signatures of the known ones.

With the aim of reducing the dependence on manually-annotated lexical resources, other research focused on modelling metaphor using corpus-driven information alone. \newcite{ShutovaColing2010} pointed out that the metaphorical uses of words constitute a large portion of the dependency features extracted for abstract concepts from corpora. For example, the feature vector for \textit{politics} would contain \textsc{game} or \textsc{mechanism} terms among the frequent features. As a result, distributional clustering of abstract nouns with such features identifies groups of diverse concepts metaphorically associated with the same source domain. \newcite{ShutovaColing2010} exploit this property of co-occurrence vectors to identify new metaphorical mappings starting from a set of examples.  \newcite{ShutovaNAACL2013} used hierarchical clustering to derive a network of concepts in which metaphorical associations are learned in an unsupervised way. 
\newcite{DoDinh2016} investigated metaphors through the task of sequence labelling, detecting metaphor related words in context.
\newcite{Gutierrez2010} investigated metaphorical composition
in the compositional distributional semantics framework. Their method learns metaphors as linear transformations
in a vector space and they demonstrated that it produces superior phrase representations for both metaphorical and literal language, as compared to the traditional "single-sense" compositional distributional model. They then used
these representations in the metaphor identification
task, achieving promising results.

The more recent approaches of \newcite{Shutova2016} and \newcite{Bulat2017} used dense skip-gram word embeddings \cite{Mikolov2013a} instead of the sparse distributional features. 
 \newcite{Shutova2016} investigated a set of metaphor identification methods using linguistic and visual features. They learned linguistic and visual representations for both words and phrases, using skip-gram and convolutional neural networks \cite{Kiela:2014} respectively. They then measured the difference between the phrase representation and those of its component words in terms of their cosine similarity, which served as a predictor of metaphoricity. 
 They found basic cosine similarity between the component words in the phrase to be a powerful measure -- the neural embeddings of the words were compared with cosine similarity and a threshold was tuned on the development set to distinguish between literal and metaphorical phrases.
This approach was their best performing linguistic model, outperformed only by a multimodal system which included both linguistic and visual features.

\newcite{Bulat2017} presented a metaphor identification method that uses representations constructed from human property norms \cite{McRae2005}. They first learn a mapping from the skip-gram embedding vector space to the property norm space using linear regression, which allows them to generate property norm representations for unseen words. The authors then train an SVM classifier to detect metaphors using these representations as input. \newcite{Bulat2017} have shown that the cognitively-driven property norms outperform standard skip-gram representations in this task.


\section{Supervised Similarity Network}
\label{sec:ssn}

Our method is inspired by the findings of \newcite{Shutova2016}, who showed that the cosine similarity between neural embeddings of the two words in a phrase is indicative of its metaphoricity. For example, the phrase \textit{`colourful personality'} receives a score:

\begin{equation}
s = cos(x_c, x_p)
\end{equation}

\noindent where $x_c$ is the embedding for \textit{colourful} and $x_p$ is the embedding for \textit{personality}. The combined phrase is classified as being metaphorical based on a threshold, which is optimised on a development dataset.
In this paper, we propose several extensions to this general idea, creating a supervised version of the cosine similarity metric which can be optimised on training data to be more suitable for metaphor detection.

\subsection{Word Representation Gating}
 Directly comparing the vector representations of both words treats each of the embeddings as an independent unit. In reality, however, word meanings vary and adapt based on the context. In case of metaphorical language (e.g. ``\textit{cure} crime''), the source domain properties of the verb (e.g. \textit{cure}) are projected onto the target domain noun (e.g. \textit{crime}), resulting in the interaction of the two domains in the interpretation of the metaphor.

In order to integrate this idea into the metaphor detection method, we can construct a gating function that modulates the representation of one word based on the other. Given embeddings $x_1$ and $x_2$, the gating values are predicted as a non-linear transformation of $x_1$ and applied to $x_2$ through element-wise multiplication:

\begin{equation}
g = \sigma(W_g x_1)
\end{equation}

\begin{equation}
\widetilde{x}_2 = x_2 \odot g
\end{equation}

\noindent where $W_g$ is a weight matrix that is optimised during training, $\sigma$ is the sigmoid activation function, and $\odot$ represents element-wise multiplication.
In an adjective-noun phrase, this architecture allows the network to first look at the adjective, then use its meaning to change the representation of the noun. The sigmoid activation function makes it act as a filter, choosing which information from the original embedding gets through to the rest of the network. While learning a more complex gating function could be beneficial for very large training resources, the filtering approach is more suitable for the annotated metaphor datasets which are relatively small in size.

\subsection{Vector Space Mapping} 

As the next step, we implement position-specific mappings for the word embeddings.
The original method uses word embeddings that have been pre-trained using the distributional skip-gram objective \cite{Mikolov2013a}. While this tunes the vectors for predicting context words, there is no reason to believe that the same space is also optimal for the task of metaphor detection. 
In order to address this shortcoming, we allow the model to learn a mapping from the skip-gram vector space to a new metaphor-specific vector space:

\begin{equation}
z_1 = tanh(W_{z_1} x_1)
\end{equation}
\vspace{-0.3cm}
\begin{equation}
z_2 = tanh(W_{z_2} \widetilde{x}_2)
\end{equation}

\noindent where $W_{z_1}$ and $W_{z_2}$ are weight matrices, $z_1$ and $z_2$ are the new position-specific word representations.
While the original embeddings $x_1$ and $x_2$ are pre-trained on a large unannotated corpus, the transformation process is optimised using annotated metaphor examples, resulting in word representations that are more suitable for this task. Furthermore, the adjectives and nouns use separate mapping weights, which allows the model to better distinguish between the different functionalities of these words. In contrast, the original cosine similarity is not position-specific and would give the same result regardless of the word order.

\subsection{Weighted Cosine}

If the vectors $x_1$ and $x_2$ are normalised to unit length, the cosine similarity between them is equal to their dot product, which in turn is equal to their elementwise multiplication followed by a sum over all elements:

\begin{equation}
cos(x_1, x_2) \propto \sum_i x_{1,i} x_{2,i}
\end{equation}

\noindent This calculation of cosine similarity can be formulated as a small neural network where the two unit-normalised input vectors are directly multiplied together. This is followed by a single output neuron, with all the intermediate weights set to value $1$. Such a network would calculate the same sum over the element-wise multiplication, outputting the value of cosine similarity.

Since there is no reason to assume that all the embedding dimensions are equally important when detecting metaphors, we can explore other strategies for weighting the similarity calculation.
\newcite{Rei2014} used a fixed formula to calculate weights for different dimensions of cosine similarity and showed that it helped in recovering hyponym relations.
We extend this even further and allow the network to use multiple different weighting strategies which are all optimised during training. This is done by first creating a vector $m$, which is an element-wise multiplication of the two word representations:

\begin{equation}
m_i = z_{1,i} z_{2,i}
\end{equation}

\noindent where $m_i$ is the $i$-th element of vector $m$ and $z_{1,i}$ is the $i$-th element of vector $z_{1}$. After that, the resulting vector is used as input for a hidden neural layer:

\begin{equation}
d = \gamma(W_d m)
\end{equation}

\noindent where $W_d$ is a weight matrix and $\gamma$ is an activation function. If the length of $d$ is $1$, all the weights in $W_d$ have value $1$, and $\gamma$ is a linear activation, then this formula is equivalent to a regular cosine similarity. However, we use a larger length for $d$ to capture more features, use $tanh$ as the activation function, and optimise the weights of $W_d$ during training, giving the framework more flexibility to customise the model for the task of metaphor detection.

\begin{table}[t]
\setlength\tabcolsep{14.5pt}
\begin{tabular}{ll} \toprule
\textbf{Metaphorical} & \textbf{Literal} \\
\hline
absorb cost & accommodate guest \\
attack problem & attack village \\
attack cancer & blur vision \\
breathe life & breathe person \\
design excuse &  deflate mattress \\
deflate economy & digest milk \\
leak news & land airplane \\
swallow anger & swim man \\\bottomrule
\end{tabular}
\caption{Annotated verb-direct object and verb-subject pairs from \textsc{moh}.}
\label{fig:MOH}
\end{table}

\subsection{Prediction and Optimisation}

Based on vector $d$ we can output a prediction for the word pair, showing whether it is literal or metaphorical:

\begin{equation}
y = \sigma(W_y d)
\end{equation}

\noindent where $W_y$ is a weight matrix, $\sigma$ is the logistic activation function, and $y$ is a real-valued prediction with values between $0$ and $1$.

We optimise the model based on an annotated training dataset, while minimising the following hinge loss function:

\begin{equation}
E = \sum_k q_k
\end{equation}
\vspace{-0.3cm}

\begin{equation}
q_k = 
\begin{cases}
(\widetilde{y} - y)^2 & \text{if } |\widetilde{y} - y| > 0.4\\
0,              & \text{otherwise}
\end{cases}
\label{eq:hinge}
\end{equation}

\noindent where $y$ is the predicted value, $\widetilde{y}$ is the true label, and $k$ iterates over all training examples. Equation \ref{eq:hinge} optimises the model to minimise the squared error between the predicted and true labels. However, this is only done for training examples where the predicted error is not already close enough to the desired result. The condition $|\widetilde{y} - y| > 0.4$ only updates training examples where the difference from the true label is greater than $0.4$. The true labels $\widetilde{y}$ can only take values $0$ (literal) or $1$ (metaphorical), and the threshold $0.4$ is chosen so that datapoints that are on the correct side of the decision boundary by more than $0.1$ would be ignored, which helps reduce overfitting and allows the model to focus on the misclassified examples. 

The diagram of the complete network can be seen in Figure \ref{fig:network}.

\section{Word Representations}\label{vectors}

Following \newcite{Bulat2017} we experiment with two types of semantic vectors: skip-gram word embeddings and attribute-based representations. 

The word embeddings are 100-dimensional and were trained using the standard log-linear skip-gram model with negative sampling of \newcite{Mikolov:2013} on Wikipedia for 3 epochs, using a symmetric window of 5 and 10 negative samples per word-context pair. 

We use the 2526-dimensional attribute-based vectors trained by \newcite{Bulat2017}, following \newcite{Fagarasan:2015}. These representations were induced by using partial least squares regression to learn a cross-modal mapping function between the word embeddings described above and the \newcite{McRae2005} property-norm semantic space.

\section{Datasets}

We evaluate our method using two datasets of phrases manually annotated for metaphoricity. Since these datasets include examples for different senses (both metaphorical and literal) of the same verbs or adjectives, they allow us to test the extent to which our model is able to discriminate between different word senses, as opposed to merely selecting the most frequent class for a given word.

\vspace{+0.2cm}
\noindent \textbf{Mohammad et al. dataset (\textsc{moh})} \hspace{+0.2cm} \newcite{Mohammad2016} used WordNet to find verbs that had between three and ten senses and extracted the sentences exemplifying them in the corresponding glosses, yielding a total of 1639 verb uses in sentences. Each of these was annotated for metaphoricity by 10 annotators via the crowd-sourcing platform CrowdFlower\footnote{www.crowdflower.com}. Mohammad et al. selected the verbs that were tagged by at least 70\% of the annotators as metaphorical or literal to create their dataset.  We extracted verb--direct object and verb--subject relations of the annotated verbs from this dataset, discarding the instances with pronominal or clausal subject or object. This resulted in a dataset of 647 verb--noun pairs  (316 metaphorical and 331 literal). Some examples of annotated verb phrases from \textsc{moh} are presented in Table~\ref{fig:MOH}.


\begin{table}[t]
\setlength\tabcolsep{16.5pt}
\begin{tabular}{ll} \toprule
\textbf{Metaphorical} & \textbf{Literal} \\
\hline
bloody stupidity &	bloody nose	\\
deep understanding	&	cold		weather	\\
empty promise	&	dry skin	\\
green energy	&	empty can	\\
healthy balance & frosty morning\\
hot topix & hot chocolate \\
muddy thinking &  gold coin \\
ripe age & soft leather \\
sour mood &  sour cherry \\
warm reception & steep hill \\\bottomrule
\end{tabular}
\caption{Annotated adjective--noun pairs from \textsc{tsv-test}.}
\label{fig:TSV}
\end{table}

\vspace{+0.2cm}
\noindent \textbf{Tsvetkov et al. dataset (\textsc{tsv})} \hspace{+0.2cm} \newcite{Boytsov2014} construct a dataset of adjective--noun pairs annotated for metaphoricity. This is divided into a training set consisting of 884 literal and 884 metaphorical pairs (\textsc{tsv-train}) and a test set containing  100 literal and 100 metaphorical pairs (\textsc{tsv-test}). Table~\ref{fig:TSV} shows a portion of annotated adjective-noun phrases from \textsc{tsv-test}.
\textsc{tsv-train} was collected from publicly available metaphor collections on the web and manually curated by removing duplicates and metaphorical phrases that depend on wider context for their interpretation (e.g. \textit{drowning students}). \textsc{tsv-test} was constructed by extracting nouns that co-occur with a list of 1000 frequent adjectives in the TenTen Web Corpus\footnote{https://www.sketchengine.co.uk/ententen-corpus/} using SketchEngine. The selected adjective-noun pairs were annotated for metaphoricity by 5 annotators with an inter-annotator agreement of $\kappa=0.76$.  
Since \textsc{tsv-train} and \textsc{tsv-test} were constructed differently, we follow previous work \cite{Boytsov2014,Shutova2016,Bulat2017} and report performance on \textsc{tsv-test}. We randomly separated 200 (out of the 1536) examples from the training set to use for development experiments.

\section{Experiments and Results}
\label{sec:eval}


The word representations in our model were initialised with either the 100-dimensional skip-gram embeddings or the 2,526-dimensional attribute vectors (Section~\ref{vectors}). These were kept fixed and not updated, which reduces overfitting on the available training examples. For both word representations we use the same embeddings as \newcite{Bulat2017}, which makes the results directly comparable and shows that the improvements are coming from the novel architecture and are not due to a different embedding initialisation.

The network was optimised using AdaDelta \cite{Zeiler2012} for controlling adaptive learning rates. 
The models were evaluated after each full pass over the training data and training was stopped if the F-score on the development set had not improved for 5 epochs.
The transformed embeddings $z_1$ and $z_2$ were set to size 300, layer $d$ was set to size 50. 
The values for these hyperparameters were chosen experimentally using the development dataset.
In order to avoid drawing conclusions based on outlier results due to random initialisations, we ran each experiment 25 times with random seeds and present the averaged results in this paper.
We implemented the framework using Theano \cite{Al-Rfou2016} and are making the source code publicly available.\footnote{http://www.marekrei.com/projects/ssn}

\begin{table}[t]
\setlength\tabcolsep{4.5pt}
\begin{tabular}{lrrrr} \toprule
 & Acc & P  & R & F1\\\midrule
Tsvetkov et al. (2014) & - & - & - & \textbf{85}\\
Shutova et al. (2016) & & & & \\
\hspace{0.4cm} linguistic & - & 73 & 80 & 76\\
\hspace{0.4cm} multimodal & - & 67 & \textbf{96} & 79\\
Bulat et al. (2017) & - & 85 & 71 & 77\\\midrule
FFN skip-gram & 77.6 & 86.6 & 65.4 & 74.4\\
FFN attribute & 76.6 & 82.0 & 68.6 & 74.5\\
SSN skip-gram & 82.2 & \textbf{91.1} & 71.6 & 80.1\\
SSN attribute & 81.9 & 86.6 & \textbf{75.7} & 80.6\\
SSN fusion & \textbf{82.9} & 90.3 & 73.8 & \textbf{81.1}\\\bottomrule
\end{tabular}
\caption{System performance on the Tsvetkov et al. dataset
(\textsc{TSV}) in terms of accuracy (Acc), precision (P), recall (R) and F-score (F1).}
\label{tab:tsvetkov}
\end{table}

Table \ref{tab:tsvetkov} contains results of different system configurations on the TSV dataset. The original F-score by \newcite{Boytsov2014} is still the highest, as they used a range of highly-engineered features that require manual annotation, such as the lexical abstractness, imageability scores and the relative number of supersenses for each word in the dataset. 
Our setup is more similar to the linguistic experiments by \newcite{Shutova2016}, where metaphor detection is performed using pretrained word embeddings. 
They also proposed combining the linguistic model with a system using visual word representations and achieved performance improvements.
Recently, \newcite{Bulat2017} compared different types of embeddings and showed that attribute-based representations can outperform regular skip-gram embeddings.

As an additional baseline, we report the performance on metaphor detection using a basic feedforward network (FFN). In this configuration, the word embeddings $x_1$ and $x_2$ are directly connected to the hidden layer $d$, skipping all the intermediate network structure. The FFN achieves $74.4\%$ F-score on \textsc{tsv-test}, showing that even such a simple model can perform relatively well in a supervised setting. Using attribute vectors instead of skip-gram embeddings gives a slight improvement, especially on the recall metric, which is consistent with the findings by \newcite{Bulat2017}.

\begin{table}[t]
\begin{tabular}{lrrrr} \toprule
 & Acc & P  & R & F1\\\midrule
Shutova et al. (2016) & & & & \\
\hspace{0.4cm} linguistic & - & 67 & 76 & 71\\
\hspace{0.4cm} multimodal & - & 65 & \textbf{87} & \textbf{75}\\\midrule
FFN skip-gram & 71.2 & 70.4 & 71.8 & 70.5\\
FFN attribute & 68.5 & 66.7 & 71.0 & 68.3\\
SSN skip-gram & \textbf{74.8} & \textbf{73.6} & \textbf{76.1} & \textbf{74.2}\\
SSN attribute & 69.7 & 68.8 & 69.7 & 68.8\\
SSN fusion & 70.8 & 70.1 & 70.9 & 69.9\\\bottomrule
\end{tabular}
\caption{System performance on the Mohammad et al. dataset
(\textsc{moh}) in terms of accuracy (Acc), precision (P), recall (R) and F-score (F1).}
\label{tab:mohammad}
\end{table}

The architecture described in Section \ref{sec:ssn}, which we refer to as a supervised similarity network (SSN), outperforms the baseline and achieves $80.1\%$ F-score using skip-gram embeddings and $80.6\%$ with attribute-based representations.
We also created a fusion of these two models where the predictions from both are combined as a weighted average. In this setting, the two networks are trained in tandem and a real-valued weight, which is also optimised during training, is used to combine them together. This configuration achieves $81.1\%$ F-score, indicating that the the skip-gram embeddings and attribute vectors capture somewhat complementary information. Excluding the system by \newcite{Boytsov2014} which requires hand-annotated features, the proposed similarity network outperforms all the previous systems, even improving over the multimodal system by \newcite{Shutova2016} without requiring any visual information. The attribute-based SSN also improves over \newcite{Bulat2017} by $5.6\%$ absolute, using the same word representations as input.

Table \ref{tab:mohammad} contains results of different system architectures on the \textsc{moh} dataset. \newcite{Shutova2016} reported $75\%$ F-score on this dataset with a multimodal system, after randomly separating a subset for testing. Since this corpus contains only 647 annotated examples, we instead evaluated the systems using 10-fold cross-validation. The feedforward baseline with skip-gram embeddings returns an F-score that is close to the linguistic configuration of Shutova et al, whereas the best results are achieved by the similarity network with skip-gram embeddings. In this setting, the attribute-based representations did not improve performance -- this is expected, as the attribute norms by \newcite{McRae2005} are designed for nouns, whereas the \textsc{moh} dataset is centered on verbs. 

Table \ref{tab:examples} contains examples from the \textsc{tsv} development set, together with gold annotations and predicted scores. The system confidently detects literal phrases such as \textit{sunny country} and \textit{meaningless discussion}, along with metaphorical phrases such as \textit{unforgiving heights} and \textit{blind hope}. The predicted output disagrees with the annotation on cases such as \textit{humane treatment} and \textit{rich programmer} -- some of these examples could also be argued as being metaphorical, depending on the specific sense of the words. While the system was relatively unsure about the false positives (the scores were close to $0.5$), it tended to assign more decisive scores to the false negatives.

\begin{table}[t]
\setlength\tabcolsep{8.0pt}
{\small
\begin{tabular}{lrrr} \toprule
Input phrase & Gold & Predicted & Score \\\midrule
sunny country & 0 & 0 & 0.152\\
sweet treat & 0 & 0 & 0.358\\
lost wallet & 0 & 0 & 0.439\\
meaningless discussion & 0 & 0 & 0.150\\
gentle soldier & 0 & 0 & 0.175\\
unforgiving heights & 1 & 1 & 0.867\\
easy money & 1 & 1 & 0.503\\
blind hope & 1 & 1 & 0.813\\
rolling hills & 1 & 1 & 0.677\\
educational gap & 1 & 1 & 0.827\\
humane treatment & 0 & 1 & 0.617\\
democratic candidate & 0 & 1 & 0.510\\
rich programmer & 0 & 1 & 0.514\\
fishy offer & 1 & 0 & 0.290\\
backward area & 1 & 0 & 0.161\\
sweet person & 1 & 0 & 0.332\\\bottomrule
\end{tabular}
}
\caption{Examples from the Tsvetkov development set, together with the gold label, predicted label, and the predicted score from the best model.}
\label{tab:examples}
\end{table}

\section{The Effects of Training Data}

Results in Section \ref{sec:eval} show that performance on the \textsc{tsv} dataset is higher than the \textsc{moh} dataset, likely due to the former having more examples available for training. Therefore, we ran an additional experiment to investigate the effect of dataset size on the performance of metaphor detection.
\newcite{Gutierrez2010} annotated a dataset of adjective-noun phrases as being literal or metaphorical, and we are able to use this as an additional training resource. While it contains only 23 unique adjectives, the total number of phrases reaches 8,592. We remove any phrases that occur in the development or test data of \textsc{tsv}, then incrementally add the remaining examples to the \textsc{tsv} training data and evaluate on the \textsc{tsv-test}.

\begin{figure}[t]
    \centering
	\includegraphics[width=\linewidth]{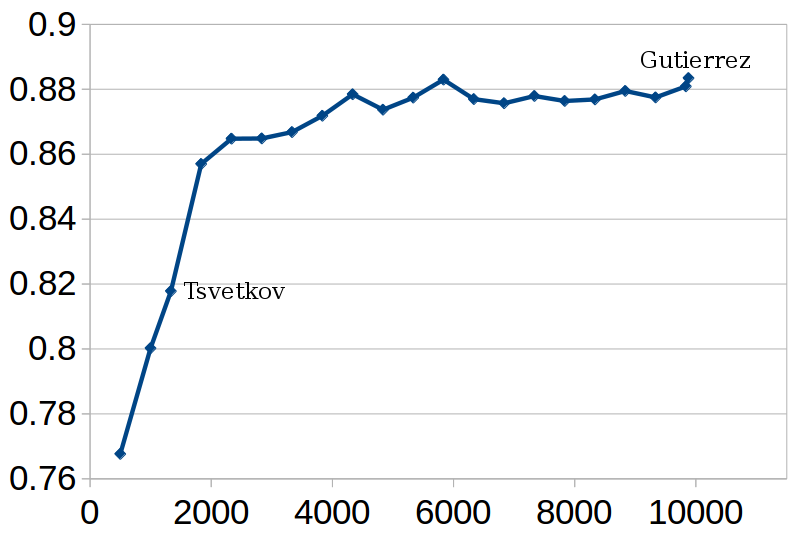}
	\caption{Performance as a function of training set size. The x-axis shows the number of training examples, the y-axis shows F-score on \textsc{tsv-test}.}
	\label{fig:moredata}
\end{figure}

Figure \ref{fig:moredata} shows a graph of the system performance, when increasing the training data at intervals of 500.
There is a very rapid increase in performance until around 2,000 training points, whereas the existing \textsc{tsv-train} is limited to 1,336 examples.
Providing even more data to the system gives an additional increase that is more gradual. The final performance of the system using both datasets is $88.3$ F-score, which is the highest result reported on the \textsc{tsv} dataset and translates to $36\%$ relative error reduction with respect to the same system trained only on the original dataset.

We report the exact values in Table \ref{tab:moredata} for the different training sets. The value on the Tsvetkov training data is different from the result in Table \ref{tab:tsvetkov}, which is due to the original attribute embeddings by \newcite{Bulat2017} only containing representations for the vocabulary in the \textsc{tsv} dataset. In order to include the data from \newcite{Gutierrez2010}, we recreated the attribute vectors for a larger vocabulary, which results in a slightly different baseline performance. 

\begin{table}[t]
\begin{tabular}{lrrrr} \toprule
Training data & Acc & P  & R & F\\\midrule
Tsvetkov & 83.0 & 88.3 & 76.3 & 81.8\\
Tsvetkov+Gutierrez & \textbf{88.7} & \textbf{91.6} & \textbf{85.4} & \textbf{88.3}\\\bottomrule
\end{tabular}
\caption{System performance on the Tsvetkov et
al. dataset (\textsc{tsv}), using additional training data.}
\label{tab:moredata}
\end{table}

\begin{figure}[th!]
    \centering
	\includegraphics[width=\linewidth]{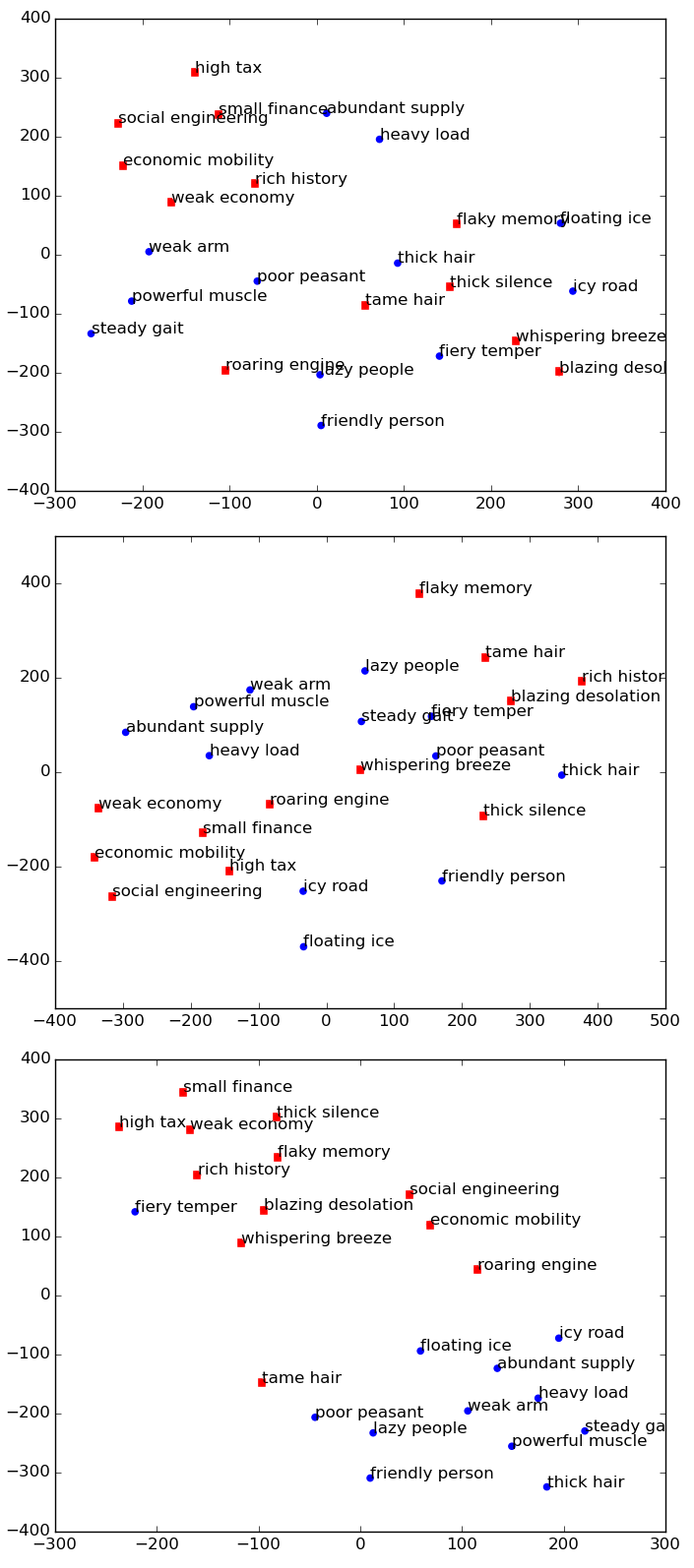}%
	\caption{Comparison of metaphorical and literal phrases in different vector spaces. Blue circles indicate literal examples, red squares show metaphorical pairs. Top: additive vector space. Middle: multiplicative vector space. Bottom: vectors from layer $m$ in the similarity network.}
	\label{fig:samplemapping}
\end{figure}

\section{Qualitative analysis}

The architecture in Section \ref{sec:ssn} also acts as a semantic composition model, extracting the meaning of the phrase by combining the meanings of its component words.
Therefore, we performed a qualitative experiment to investigate: (1) how well do traditional compositional methods capture metaphors, without any fine-tuning; and (2) whether the supervised representations still retain their domain-specific semantic information. For this purpose, we construct three vector spaces and visualise some examples from the \textsc{tsv} training set, using t-SNE \cite{VanDerMaaten2008}.

Figure \ref{fig:samplemapping} contains examples for three different composition methods: the additive method simply sums the skip-gram embeddings for both words (top); the multiplicative method multiplies the skip-gram embeddings (middle); the final system uses layer $m$ from the SSN model to represent the phrases (bottom).

The visualisation shows that the additive and multiplicative models are both comparable when it comes to semantic clustering of the phrases, but metaphorical examples are mixed together with literal clusters. The SSN is optimised for metaphor classification and therefore it produces representations with a very clear boundary for metaphoricity. Interestingly, the graph also reveals a misannotated example in the dataset, since \textit{`fiery temper'} should be labeled as a metaphor.
At the same time, this space also retains the general semantic information, as similar phrases with the same label are still positioned close together. Future work could investigate models of multi-task training where metaphor detection is trained together with an unsupervised objective, allowing the system to take better advantage of unlabeled data while still learning to separate metaphors.

\section{Conclusion}

In this paper, we introduced the first deep learning architecture designed to capture metaphorical composition and evaluated it on a metaphor identification task. 

Firstly, we demonstrated that the proposed framework outperforms both a metaphor-agnostic baseline (a feed-forward neural network) as well as previous corpus-driven approaches to metaphor identification. 
The results showed that it is beneficial to construct a specialised network architecture for metaphor detection, which includes a gating function for capturing the interaction between the source and target domains, word embeddings mapped to a metaphor-specific space, and optimisation using a hinge loss function. 

Secondly, our qualitative analysis indicates that our supervised similarity network learns phrase representations with a very clear boundary for metaphoricity, in contrast to traditional compositional methods. 

Finally, we show that with a sufficiently large training set our model can also outperform the state-of-the art metaphor identification systems based on hand-coded lexical knowledge. 

\section*{Acknowledgments}

Ekaterina Shutova's research is supported by the Leverhulme Trust Early Career Fellowship.

\bibliography{refs}
\bibliographystyle{emnlp_natbib}

\end{document}